%% file: revolver.tex
\documentclass{article}

\usepackage[preprint,nonatbib]{nips_2018}

\usepackage[sort, numbers, square]{natbib}

\usepackage[utf8]{inputenc} % allow utf-8 input
\usepackage[T1]{fontenc}    % use 8-bit T1 fonts
\usepackage{hyperref}       % hyperlinks
\usepackage{url}            % simple URL typesetting
\usepackage{booktabs}       % professional-quality tables
\usepackage{amsfonts,amsmath,amssymb}  %blackboard symbols, math environments, and more symbols
\usepackage{nicefrac}       % compact symbols for 1/2, etc.
\usepackage{microtype}      % microtypography

\usepackage{graphicx}
\usepackage{subcaption}
\usepackage{wrapfig}

\usepackage{color}
\usepackage{colortbl}

\usepackage{eqnarray}

\newcommand{\minisection}[1]{\textbf{#1}\hspace{0.3em}}

\newcommand{\Task}{\mathcal{T}}
\newcommand{\supp}{x}
\newcommand{\qry}{\bar{x}}
\newcommand{\anno}{L}
\newcommand{\lbl}{l}
\newcommand{\Targ}{\mathcal{Y}}
\newcommand{\targ}{y}
\newcommand{\pred}{\hat{y}}
\newcommand{\pix}{p}

\newcommand{\twodots}{\mathinner {\ldotp \ldotp}}
\DeclareMathOperator*{\argmin}{arg\,min}

\definecolor{Gray}{gray}{0.9}

\setcitestyle{numbers, square}

\title{Few-Shot Segmentation Propagation\\with Guided Networks}

\author{Kate Rakelly$^*$ \quad Evan Shelhamer$^*$ \quad Trevor Darrell \quad Alexei Efros \quad Sergey Levine \\
UC Berkeley\\
{\tt\small \{rakelly,shelhamer,efros,slevine,trevor\}@cs.berkeley.edu}
}

\begin{document}

\maketitle

\begin{abstract}
Learning-based methods for visual segmentation have made progress on particular types of segmentation tasks, but are limited by the necessary supervision, the narrow definitions of fixed tasks, and the lack of control during inference for correcting errors.
To remedy the rigidity and annotation burden of standard approaches, we address the problem of few-shot segmentation: given few image and few pixel supervision, segment any images accordingly.
We propose guided networks, which extract a latent task representation from any amount of supervision, and optimize our architecture end-to-end for fast, accurate few-shot segmentation.
Our method can switch tasks without further optimization and quickly update when given more guidance.
We report the first results for segmentation from one pixel per concept and show real-time interactive video segmentation.
Our unified approach propagates pixel annotations across space for interactive segmentation, across time for video segmentation, and across scenes for semantic segmentation.
Our guided segmentor is state-of-the-art in accuracy for the amount of annotation and time.
See http://github.com/shelhamer/revolver for code, models, and more details.
\end{abstract}

\input{intro}
\input{related_work}
\input{few_shot}
\input{guidance}
\input{results}

\input{discussion}

%\subsubsection*{Acknowledgments}

\bibliography{revolver}
\bibliographystyle{icml2017}

\end{document}

%% file: intro.tex
\section{Introduction}
% - we do inference on a different image than we annotate
% - we do sparse annotations
% - we are fast
% - we are correctable

Learning a particular type of segmentation, or even extending an existing model to a new task like a new semantic class, generally requires collecting and annotating a large amount of data and (re-)training the model for many iterations.
Current methods are supervised by fully annotated images numbering in the thousands or tens of thousands, such that even a ``small'' dataset contains billions of pixelwise annotations.  %% thousands \citep{pascal,sbdd}, 100k \citep{coco}
Collecting these dense annotations is time-consuming, tedious, and error-prone.
There are many tasks of practical and scientific interest for which annotation on this scale is impractical or even infeasible, such as graphic design, medical imaging, and more.

Semi- and weakly-supervised segmentation methods can propagate annotations across inputs within a task (instance segmentation throughout a video) or across different types of annotations (tags, boxes, and masks), but current approaches are specific to tasks or forms of supervision, and are often inefficient in computation or data.
Once learned, these methods are difficult to guide or correct, and are insensitive to small amounts of further annotation.
On the other hand, interactive segmentation methods adjust to a given task with few annotations, and can be corrected.
However, annotations only control inference on that same image and cannot inform the segmentation of a new input.

\begin{figure}
\centering
\includegraphics[width=\textwidth]{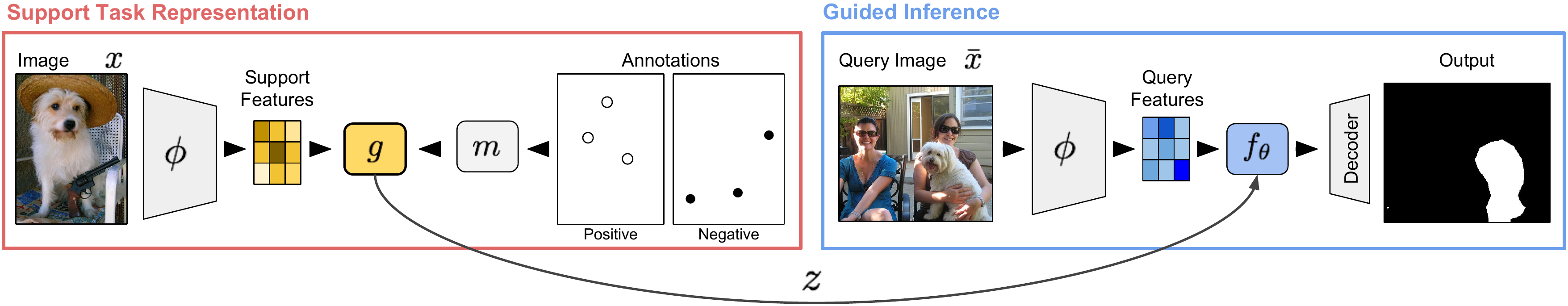}
\caption{
Guided network for few-shot segmentation.
See Section \ref{sec:guided} for details.
}
\label{fig:model}
\end{figure}

We instead address the problem of few-shot segmentation: given only a few images with sparse pixelwise annotations that indicate the task, segment unannotated images accordingly.
Our unifying framework is ``pixels in, pixels out,'' for propagating any collection of pixel annotations, from within and across images, to unannotated pixels for inference.
We directly optimize a guided network to infer a latent task defined by sparse annotations and segment new inputs conditioned on that task.
Our few-shot segmentor segments new concepts from as little as one pixel per concept and incorporates further annotations near-instantaneously to update and improve inference.
Existing methods, designed for specific segmentation tasks, fail in the extremely sparse regime while our method can propagate across the spectrum from one annotated pixel up to full, dense masks.
Our few-shot segmentor is task-agnostic in switching as directed by annotations, data efficient in learning from few pixelwise annotations, and correctable in incrementally incorporating more supervision.

The few-shot setting divides the input into an annotated support, which supervises the task to be done, and an unannotated query that should be segmented accordingly.
In this work we address these key parts of the few-shot segmentation problem: (1) how to summarize the sparse, structured support into a task representation, (2) how to condition pixelwise inference on the given task representation, and (3) how to synthesize segmentation tasks for accuracy and generality.
Structured output poses challenges for each of these aspects due to its high-dimensional, statistically dependent, and skewed input and output distributions.
We make connections to few-shot methods in the image classification setting as we adapt them to segmentation for comparison with our approach.

We propose a new class of \emph{guided} networks that extend few-shot and fully convolutional architectures; see Figure~\ref{fig:model}.
Given an annotated support set and query image, the guide $g$ extracts a latent representation $z$ of the task, which directs the segmentation of the query by $f_\theta$.
We carry out a comprehensive comparison of how to encode the support (Section \ref{sec:task-repr}), and introduce a new mechanism for fusing images and annotations that improves both learning time and inference accuracy.
We examine different choices of guided inference (Section \ref{sec:co-inference}) to identify which is best for structured output.
Once trained, our model requires no further optimization to handle new few-shot tasks, and can swiftly and incrementally incorporate additional annotations to alter the task or correct errors.

We evaluate our method on a variety of challenging segmentation problems: interactive image segmentation in \ref{sec:res-inter}, semantic segmentation in \ref{sec:res-sem}, video object segmentation in \ref{sec:res-vos}, and real-time interactive video segmentation in \ref{sec:res-iv}.
See Figure \ref{fig:tasks} for an illustration of the problems we consider.
The focus of our results is in the sparse regime, for which it is practical to collect annotations.
In all cases our accuracy is state-of-the-art for the amount of annotations and time required.
The speed with which our method incorporates new annotations makes it suitable for real-time interactive use.

%% file: related_work.tex
\section{Related Work}

Our framework extends and bridges segmentation and few-shot learning.
Segmentation is a vast subject with many current directions for deep learning techniques \cite{garcia2017review}.
We focus on one-shot, semi-supervised, and interactive methods, which are addressed separately in existing work.
We review few-shot learning methods and how they relate to structured output.

\begin{figure}[t]
\centering
\includegraphics[width=\textwidth]{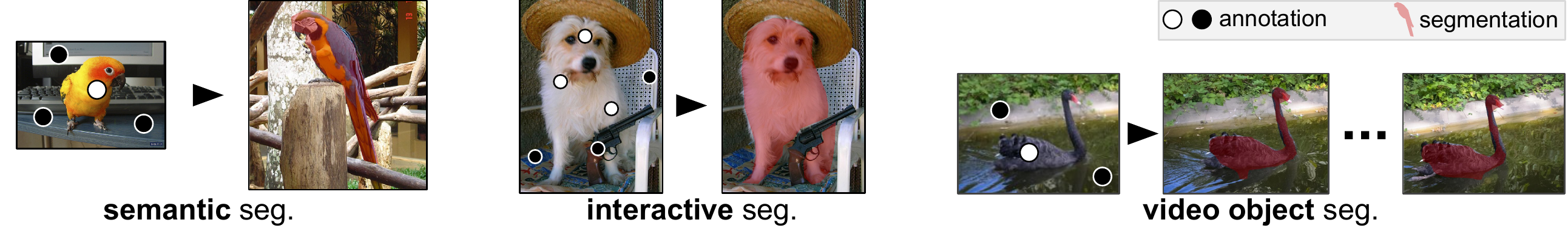}
\caption{
Few-shot segmentation subsumes semantic, interactive, and video object segmentation.
}
\label{fig:tasks}
\end{figure}

\minisection{Segmentation} There are many modes of segmentation.
We take up semantic \cite{pascal,liu2011sift}, interactive \cite{kass1988snakes,boykov2001interactive}, and video object segmentation \cite{pont-tuset2017davis} as our challenge problems.
See Fig. \ref{fig:tasks} for summary.

For semantic seg., \citet{shaban2017one} proposes a pioneering but limited one-shot segmentor (OSLSM), which requires few images but dense annotations and needs semantic supervision at training time.
Our few-shot segmentor can segment a class from as few as two points for positive and negative, and even few-shot segment classes from instance supervision, due to our guided architecture.
For video object seg., one-shot video object segmentation (OSVOS) by \citet{caelles2016one} achieve high accuracy by fine-tuning during inference, but this optimization is too costly in time and fails with sparse annotations.
By contrast, we learn to guide segmentation propagation from sparse annotations by synthesizing tasks with that kind of annotation.
We rely on feedforward guidance to few-shot learn, making our method much faster.
For interactive seg., \citet{xu2016deep} learn state-of-the-art interactive object segmentation (DIOS), but cannot propagate annotations across different images as we do.
This is a bottleneck on annotation efficiency, since it requires $>=2$ annotations for every input, while our method can segment new inputs on its own.
Note that interactive is a special case of few-shot.

\minisection{Few-shot learning} Few-shot learning \cite{fei2006one,lake2015human} holds the promise of data efficiency: in the extreme case, one-shot learning requires only a single annotation of a new concept.
The present wave of methods \cite{koch2015siamese,santoro2016meta,vinyals2016matching,finn2017model,ravi2017optimization,snell2017prototypical} frame it as direct optimization for the few-shot setting: they synthesize few-shot tasks, define a task loss, and learn a task model given the few-shot supervision.
Existing works focus on classification, not structured output, and have given little attention to sparse and imbalanced supervision.
We are the first to address general few-shot segmentation, ranging from extremely sparse to dense annotations, with efficiency for use on image collections and video.

To situate our work, we summarize approaches as they relate to visual structured output tasks like segmentation.
Few-shot as optimization \cite{ravi2017optimization,finn2017model} optimizes during inference by gradients on pre-training for fine-tuneability (MAML) \cite{finn2017model} or a learned recurrent optimizer \cite{ravi2017optimization}.
While remarkable in generality by lack of task and architecture assumptions, the second-order optimization and learned optimization of these methods are computationally impractical on large-scale networks for visual recognition, and are unproven for the high dimensionality and skewed distributions of segmentation.
Few-shot as embedding \cite{koch2015siamese,santoro2016meta,vinyals2016matching,snell2017prototypical} learns a metric and retrieves the closest item from the support, inspired by siamese networks for metric learning \cite{chopra2005learning,hadsell2006dimensionality}.
These methods are fast on small problems, and can be admirably simple \cite{snell2017prototypical}, but degrade with higher shot and way.
Structured output tasks can have $~10^6$ elements in one support and query, and the full way of instance recognition can be in the $1000+$.
Few-shot as modulation \cite{wang2016learning,bertinetto2016learning} regresses task model parameters based on the support.
While adopted by OSLSM, it is difficult to merge the modulations for $>1$ shot (and they do not).

%% file: few_shot.tex
\section{Few-Shot Segmentation}
\label{sec:few-shot}

Few-shot learning divides the input into an annotated support, which supervises the task to be done, and an unannotated query on which to do the task.
The common setting in which the support contains $K$ distinct classes and $S$ examples of each is referred to as $K$-way, $S$-shot learning \citep{lake2015human,fei2006one,vinyals2016matching}.
For few-shot segmentation tasks we add a further pixel dimension to this setting, as we must now consider the number of support pixel annotations for each image, as well as the number of annotated support images.
We denote the number of annotated pixels per image as $P$, and consider the settings of $(S,P)$-shot learning for various $S$ and $P$.
In particular, we focus on sparse annotations where $P$ is small, as these are more practical to collect, and merely require the annotator to point to the segment(s) of interest.
This type of data collection is more efficient than collecting dense masks by at least an order of magnitude \citep{bearman2016point}.
As a further step, we handle mixed-shot learning where the amount of annotation varies by class and task since structured output distributions are imbalanced.

We define a few-shot segmentation task as the set of input-output pairs $(\Task_i, \Targ_i)$ sampled from a task distribution $\mathcal{P}$, adopting and extending the notation of \citet{garcia2018few}. The task inputs are
\begin{align*}
\Task   &= \left\{ \{(\supp_1, \anno_1), \dots (\supp_S, \anno_S)\}, \{\qry_1, \dots, \qry_Q\}~;~\supp_s, \qry_q \sim \mathcal{P}_l(\mathbb{R}^N) \right\} \\
\anno_s &= \left\{ (\pix_j, \lbl_j) : j \in \{1 \twodots P\} \right\}, ~ \lbl \in \{1 \twodots K\} \cup \{\varnothing\} \\
\end{align*}
where $S$ is the number of annotated support images $\supp_s$, $Q$ is the number of unannotated query images $\qry_q$, and $\anno_s$ are the support annotations.
The annotations are sets of point-label pairs $(\pix, \lbl)$ with $|\anno_s| = P$ per image, where every label $\lbl$ is one of the $K$ classes or unknown ($\varnothing$).
The task outputs, that is the targets for the support-defined segmentation task on the queries, are
\begin{align*}
\Targ   &= (\targ_1, \dots, \targ_Q), \quad \targ_q = \left\{ (\pix_j, \lbl_j) : j \in \qry_q \right\}
\end{align*}
Without loss of generality we consider every task to be binary with $K=2$, or $\anno_s = (+_s, -_s)$, where each task defines its own positive and negative is the complement (background, as it is known in segmentation).
The choice of binary tasks is a natural one for the problems of interactive segmentation and video object segmentation, in which cases the tasks consist of a single object to be segmented.
Note that higher-way tasks can be handled as a union of binary tasks.
We let $Q=1$ throughout, since inference for each query is independent in our model.

Our approach for the few-shot segmentation problem has two parts: (1) extracting a task representation from the semi-supervised, structured support and (2) segmenting the query given the task representation.
We define the task representation as $z = g(x, +, -)$, and the query segmentation guided by that representation as $\pred = f(\qry, z)$ .
The design of the task representation $z$ and its encoder $g$ is crucial for few-shot segmentation to handle
the hierarchical structure of images and pixels,
the high and variable dimensions of images and their pixelwise annotations,
the semi-supervised nature of support with many unannotated pixels,
and skewed support distributions,
which we address in Section \ref{sec:task-repr}.
For segmentation given the task representation we meld few-shot techniques to dense, pixelwise inference via fully convolutional networks.
While related to few-shot methods on simple, low-dimensional data, our evaluation is the first comparison of these methods on large-scale visual recognition problems that stress the limits of shot, task diversity, and efficiency.

%% file: guidance.tex
\section{Guided Networks}
\label{sec:guided}

Guided networks reconcile autonomous and interactive modes of inference: a ``guided'' model is both able to make predictions on its own and incorporate expert guidance for directing the task or correcting errors.
Guidance is extracted from the support as a latent task representation $z$ through a guide $g(\supp, \anno)$.
The model carries out inference in the form of $\pred = f(\qry, z)$ for inputs $\qry$, without further need for the support.
That is, $(\pred, \qry) \perp (\supp, \anno) \mid z$.
The guidance, unlike static model parameters, is not fixed: it can be extended or corrected as directed by an annotator such as a human-in-the-loop.

We examine how to best design the guide $g$ and inference $f$ functions as deep networks.
Our method is one part architecture and one part optimization.
For architecture, we define branched fully convolutional networks, with a guide branch for extracting the task representation from the support (Section \ref{sec:task-repr}), and an inference branch for segmenting queries given the guidance (Section \ref{sec:co-inference}).
For optimization, we adapt episodic training of few-shot learning to image-to-image learning for structured output (Section \ref{sec:co-opt}).

The backbone of our networks is VGG-16 \citep{vgg}, pre-trained on ILSVRC \cite{ilsvrc}, and cast into fully convolutional form \citep{shelhamer2016fully}.
The same is true for the methods we compare against in our results.

\subsection{Guidance: From Support to Task Representation}
\label{sec:task-repr}

\begin{figure}[t]
\centering
\includegraphics[width=\textwidth]{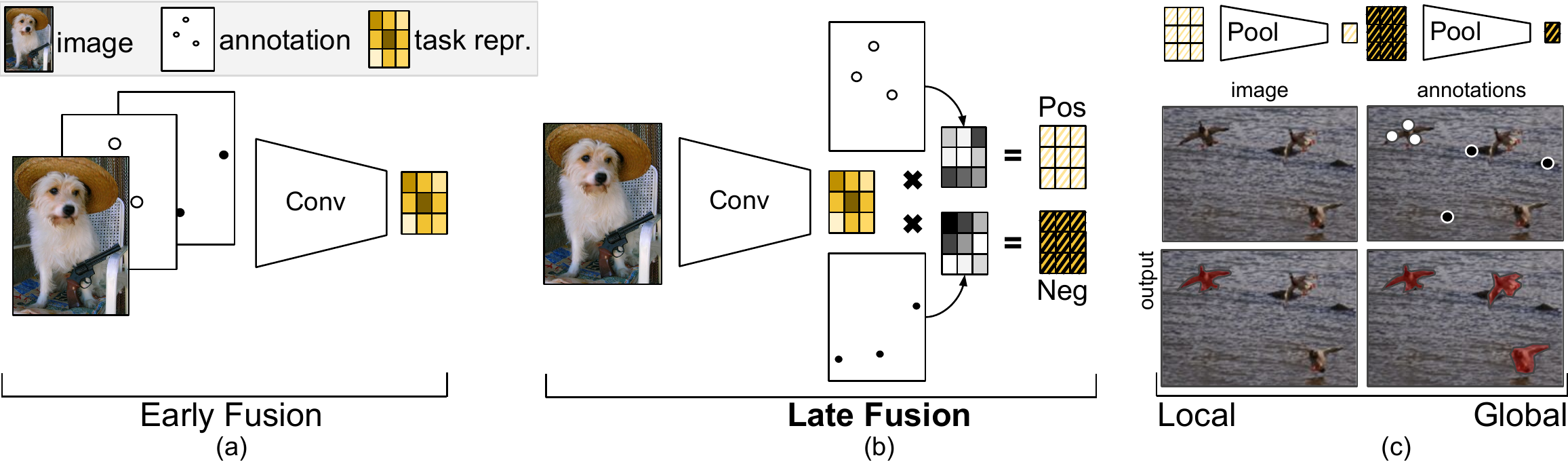}
\caption{
Extracting a task representation or ``guidance'' from the support.
(a) Early fusion simply concatenates the image and annotations.
(b) Our late fusion factorizes into image and annotation streams, improves accuracy, and updates quickly given new annotations.
(c) Globalizing the task representation propagates appearance non-locally: a single bird is annotated in this example, but global guidance causes all the similar-looking birds to be segmented (red) regardless of location.
}
\label{fig:fusion}
\end{figure}

The task representation $z$ must fuse the visual information from the image with the annotations in order to capture what should be segmented in the query.
As images with (partial) segmentations, our support is statistically dependent because pixels are spatially correlated, semi-supervised because the full supervision is arduous to annotate, and high dimensional and class-skewed because scenes are sizable and complicated.
For simplicity, we first consider a $(1, P)$-shot support consisting of one image with an arbitrary number of annotated pixels $P$, and then extend to general $(S, P)$-shot support.
As a first step we decompose the support encoder $g(x_s, +_s, -_s)$ across receptive fields indexed by $i$ for local task representations $z_i = g(x_{si}, +_{si}, -_{si})$; this is the same independence assumption made by existing fully convolutional approaches to structured output.
See Figure \ref{fig:fusion} for an overview.

\minisection{Early Fusion (prior work)}
Stacking the image and annotations channel-wise at the input makes $z_{si} = g_{\text{early}}(x, +, -) = \phi_S(x \oplus + \oplus -)$ with a support feature extractor $\phi_S$.
This early fusion strategy, employed by \cite{xu2016deep}, gives end-to-end learning full control of how to fuse.
Masking the image by the positive pixels \citep{shaban2017one,yoon2017pixel} instead forces invariance to context, potentially speeding up learning, but makes unnatural images and precludes learning from the background.
Both early fusion techniques suffer from an inherent modeling issue: incompatibility of the support and query representations.
Stacking requires distinct $\phi_S, \phi_Q$ while masking divides the input distribution.
Early fusion is slow, since changes in annotations trigger a full pass through the network.

\minisection{Late Fusion (ours)}
We resolve the learning and inference issues of early fusion by architecturally inducing structure to make $z_{si} = g_{\text{late}}(x, +, -) = \psi(\phi(\qry), m(+), m(-))$.
We first extract visual features from the image alone by $\phi(\supp)$, map the annotations into masks in the feature layer coordinates $m(+), m(-)$, and then fuse both by $\psi$ chosen to be element-wise product.
This factorization into visual and annotation branches defines the spatial relationship between image and annotations, improving data efficiency by not learning and computation time by caching.
Fixing $m$ to interpolation and $\psi$ to multiplication, the task representation can be updated quickly by only recomputing the masking and not features $\phi$.
The visual representations of the support and query are unified by design in a shared feature extractor $\phi$, improving learning efficiency.
Late fusion improves task accuracy, with $60\%$ relative improvement for video object segmentation (sec. \ref{sec:res-vos}).
Furthermore it reduces inference time, as only the masking needs to be recomputed to incorporate new annotations, making real-time interactive video segmentation possible (section \ref{sec:res-iv}).
By contrast, existing methods require seconds or even minutes to update between interactions.

At first glance, late fusion might seem limited in generality and resolution; however, we can improve these while keeping its advantages.
The annotation-to-mask mapping $m$ can be learned end-to-end to capture characteristic patterns of annotations, but more efficiently than early fusion, because it does not have to learn to mask exhaustively across visual variations, unlike the monolithic $\phi_S$.
The spatial resolution of the support can be improved by feature interpolation in $\psi$ or dilation \cite{chen2015semantic,yu2015multi} in $\phi$.
Note this can be done solely for the support without altering query inference time and output resolution.

\minisection{Locality}
We are generally interested in segmentation tasks that are determined by visual characteristics and not absolute location in space or time, i.e. the task is to group pixels of an object and not pixels in the bottom-left of an image.
When the support and query images differ, there is no known spatial correspondence, and the only mapping between support and query should be through features.
To fit the architecture to this task structure, we merge the local task representations by $m_P(\{z_{si} : \forall i\})$ for all positions $i$.
Choosing global pooling for $m_P$ globalizes the task by discarding the spatial dimensions.
The pooling step can be done by averaging, our choice, or other reductions.

However, when the support and query images are the same (e.g. interactive segmentation), feature location can be informative, and $m_P$ can instead be the identity.
The effect of this decision is seen in the segmentation of a single image with multiple objects shown by Figure \ref{fig:fusion} (right).

\minisection{Multi-Shot}
The full $(S,P)$-shot setting requires summarizing the entire support with a variable number of images with varying amounts of pixelwise annotations.
Note in this case that the annotations might be divided across the support, for instance one frame of a video may only have positives while a different frame has only negatives, so $S$-shot cannot simply be reduced to $1$-shot, as done in prior work \cite{shaban2017one}.
We form the full task representation $z_S = m_S(\{z_1, \dots, z_S\})$ by merging the shot-wise image-annotation representations $z_s$.
As long as the merge $m_S$ is differentiable the full $(S,P)$-shot task representation can be learned end-to-end.
Alternatively, optimizing for $1$-shot then merging by a fixed $m_S$ such as averaging can suffice, and requires less computation during training.

\subsection{Guiding Inference}
\label{sec:co-inference}

\begin{figure}
\centering
\includegraphics[width=\textwidth]{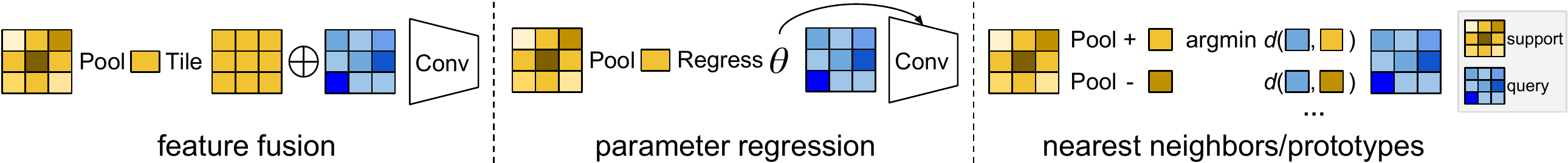}
\caption{
Types of guided inference for directing the model via the task representation.
}
\label{fig:guided-infererence}
\end{figure}

Inference in a fixed segmentation model is simply $\pred = f_\theta(\qry)$ for output $y$, parameters $\theta$, and input $\qry$.
Guided inference is a function $\pred = f(\qry, z)$ of the query conditioned on the guidance extracted from the support.
In our case inference has the further form of $f(\phi(\qry), z)$, where $\phi$ is a fully convolutional encoder from input pixels to visual features.
We explore a variety of conditioning approaches, note connections to recently proposed few-shot learning methods for image classification, and determine which most effectively copes with structured output across several types of segmentation tasks.
See Figure \ref{fig:guided-infererence} for a schematic illustration of the types of conditioning.

\minisection{Feature Fusion} $\pred = f_\theta(m_f(\phi(\qry), z))$ for fusion operation $m_f$.
In particular we consider fusing by $m_f = \phi(x) \oplus \text{tile}(z)$ which concatenates the task representation with the query features (while tiling $z$ to the spatial dimensions of the query if need be).
The fused query-support feature is then decoded to a binary segmentation by a small convolutional network $f_\theta$ that can be interpreted as a learned distance metric for retrieval from support to query.
Note that the pixel-level matching architecture of \citet{yoon2017pixel} resembles our instantiation of this approach, but without addressing multi-shot and sparse pixel settings, and furthermore they require optimization during inference for few-shot usage.

\minisection{Parameter Regression} $\pred = f_{\theta, \eta(z)}(\phi(x))$ for static inference parameters $\theta$ and dynamic task parameters regressed by $\eta(z)$.
The task parameters can correspond to any differentiable model, but in existing methods and this work they control the output layer of a fully convolutional network.
The remaining majority of the weights $\theta$ for other layers are learned by backpropagation during training then fixed.
Note that \citet{shaban2017one} regress parameters, but their regressor is global, instead of local and fully convolutional.
We find that this approach is unstable to optimize, and can sometimes converge to constant task weights that do not vary with the support.

\minisection{Nearest Neighbors and Prototypes} $\pred = \argmin_j f(\phi(\qry), z_j)$ for any distance metric $f$.
If the local task representations are kept, the query can be inferred by nearest neighbors.
Globalizing the task representation class-wise amounts to clustering fully convolutional features into prototypes.
In this case the distance from the query to the prototypes implicitly defines the task model.
This is a natural extension of \cite{snell2017prototypical} to segmentation, but we find it is difficult to optimize for structured output, perhaps due to multi-modality or fine-tuning from  pre-training.

To decide how to guide inference, we carry out pilot experiments on interactive and few-shot semantic segmentation.
In comparing feature fusion, parameter regression, and prototypes we find feature fusion is more accurate and simpler to tune.
This is the inference strategy used for all of our results.

\subsection{Optimization and Task Sampling}
\label{sec:co-opt}

\begin{figure}
\centering
\includegraphics[width=\textwidth]{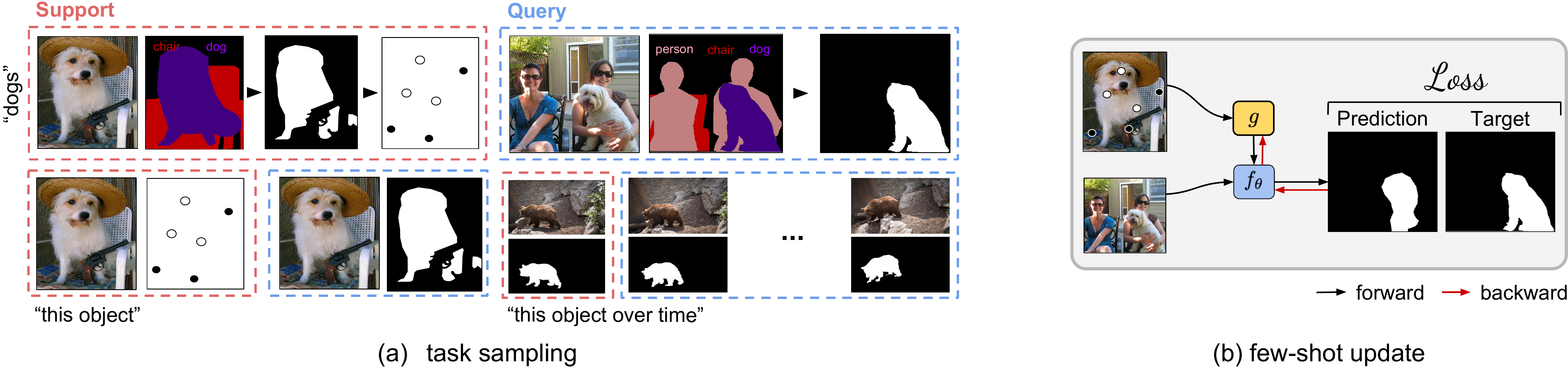}
\caption{
Optimization for few-shot segmentation propagation.
(a) Synthesizing tasks from densely annotated segmentation data.
(b) One task update: episodic training reduces to supervised learning.
}
\label{fig:few-shot-opt}
\end{figure}

In standard supervised learning, the model parameters $\theta$ are optimized according to the loss between prediction $\pred = f_\theta(x)$ and target $\targ$.
In our case, we optimize the parameters of the guide $z = g(\supp_s, +_s, -_s)$, as well as the parameters of the query segmentor $f_\theta(\qry, z)$.
We reduce the problem to supervised learning by training episodically to directly optimize for the few-shot task while learning the parameters of both branches jointly and end-to-end as shown in Figure \ref{fig:few-shot-opt}.

Inputs (annotated support and query) and (query) targets are synthesized from a fully labeled dataset.
For clarity, we distinguish between binary segmentation problems (tasks) and tasks grouped by problem distribution (modes).
For example, semantic segmentation is a mode while segmenting the category of birds is a task.
To construct a training episode, we first sample a task, then a subset of images containing that task which we divide into support and query.
We binarize support and query annotations to encode the task, and spatially sample support annotations for sparsity.
Given inputs and targets, we train the network with cross-entropy loss between the prediction and the (dense) query target.
We distinguish between optimizing the parameters of the model during training (learning) and performing few-shot learning during inference (guidance).
After learning, the model parameters are fixed, and few-shot learning is achieved via guidance and guided inference.
While we evaluate the trained network for varying support size $S$, as described in \ref{sec:co-inference}, we train with $S=1$ for efficiency.
Once learned, our guided networks can operate at different $(S, P)$ shots to address sparse and dense pixelwise annotations with the same model, unlike existing methods that train for particular shot and way.
In our experiments, we train with tasks sampled from a single mode of segmentation, but co- or cross-supervision of modes is possible.
Intriguingly, we see some transfer between modes when evaluating a guided network on a different mode than it was trained on in Section \ref{sec:res-sem}.

%% file: results.tex
\section{Results}

We evaluate our few-shot segmentor, \emph{revolver}, on a variety of problems that are representative of segmentation as a whole: interactive segmentation, semantic segmentation, and video object segmentation.
These are conventionally regarded as separate, with their own aspects and proper approaches, but we reduce each to few-shot segmentation propagation.
As a further demonstration of our framework we present results for \emph{real-time, interactive} video segmentation from dot annotations.
Every task is framed as binary segmentation of the support (+) and its complement (-).

To standardize evaluation we select one metric for all tasks: the intersection-over-union (IU) of the positives.
This is the usual choice of metric for many segmentation problems, including semantic segmentation \citep{pascal} and video object segmentation \citep{pont-tuset2017davis}, and so we adopt it here.
As a further plus, this choice allows us to compare scores across the different problems and tasks we consider.

We include fine-tuning and foreground-background segmentation as baselines for all problems.
Fine-tuning simply attempts to optimize the model on the support.
Foreground-background verifies that few-shot methods learn, and that their output co-varies with the support.
Although the tasks for few-shot segmentation vary with episodes, episodic evaluation cannot give a thorough measure of performance if the data is too simple (if there is only one object in an image for instance).
Foreground-background accuracy sets the floor for few-shot without having to alter existing benchmarks.

\input{results_interim}
\input{results_sem}
\input{results_vos}
\input{results_intervid}

%% file: results_interim.tex
\subsection{Interactive Image Segmentation}  % A.K.A. interactive object selection
\label{sec:res-inter}

We recover this problem as a special case of few-shot segmentation when the support and query images are identical.
We compare to deep interactive object selection (DIOS) \citep{xu2016deep}, because it is state-of-the-art and shares our focus on learning for label efficiency and generality.
Our approach differ in support encoding: DIOS fuses early while we fuse late and globally.
Revolver is more accurate with extreme sparsity and intrinsically faster to update, as DIOS must do a full forward pass.
See Figure \ref{fig:res-inter-sem}.
From this result we decide on late-global guidance throughout.

%% file: results_sem.tex
\begin{figure}
\centering
\begin{subfigure}{0.45\textwidth}
\centering
\includegraphics[width=0.9\textwidth]{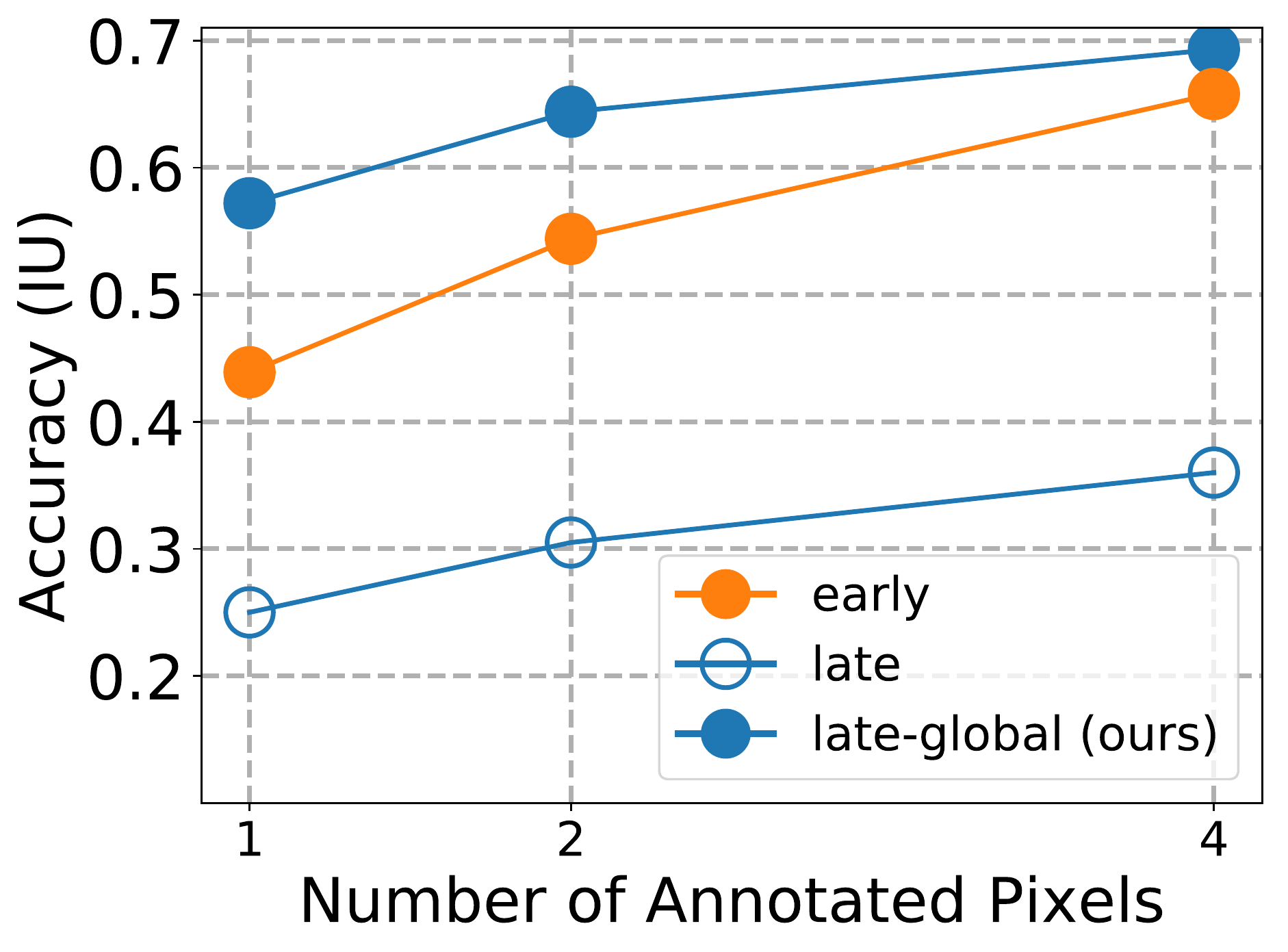}
\end{subfigure}%
\begin{subfigure}{0.45\textwidth}
\centering
\includegraphics[width=\textwidth]{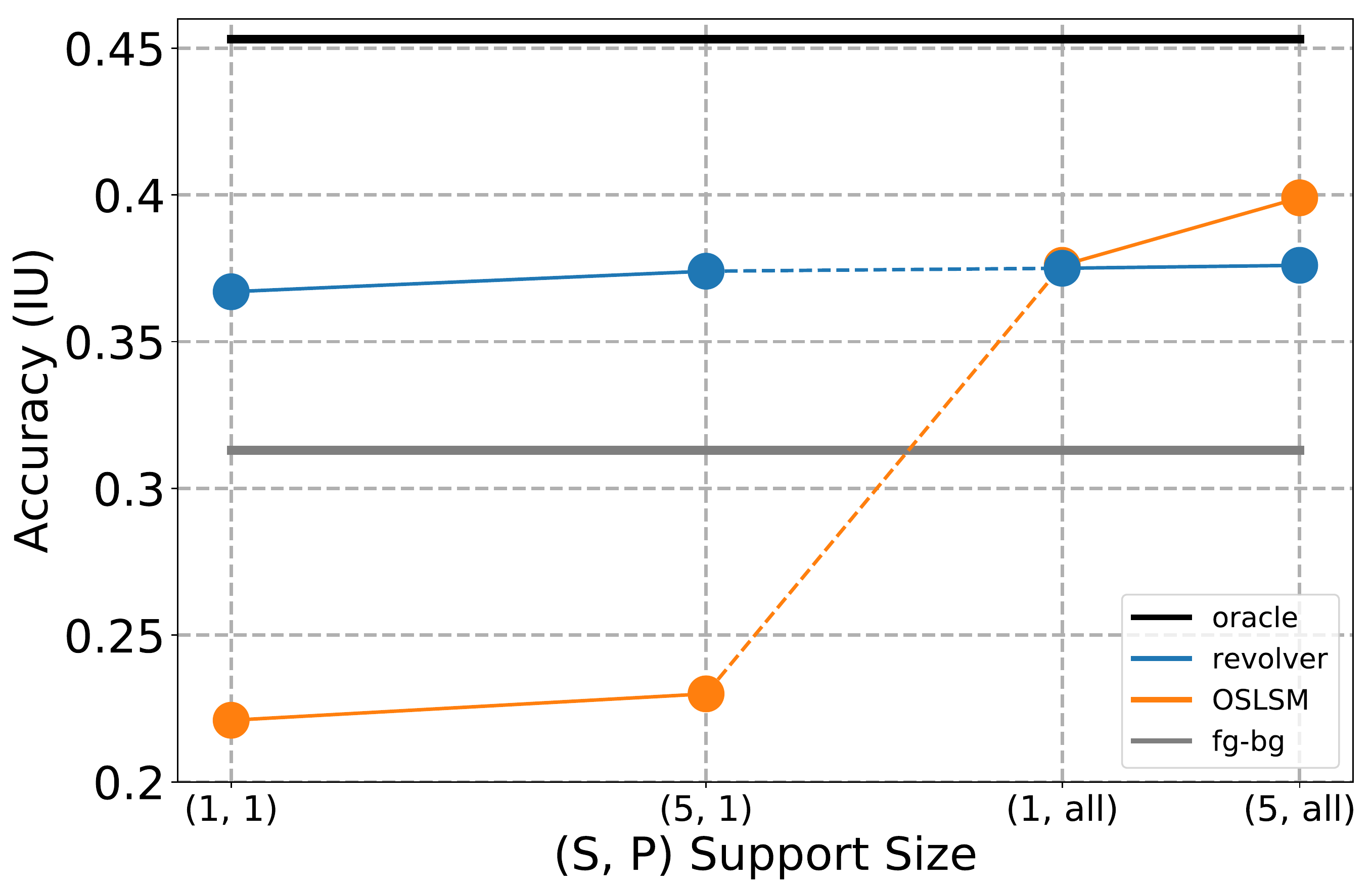}
\end{subfigure}%
\caption{
(left) Interactive segmentation of objects in images.
(right) Few-shot semantic segmentation of held-out classes: we are state-of-the-art with only two points and competitive with full annotations.
}
\label{fig:res-inter-sem}
\end{figure}

\subsection{Few-shot Semantic Segmentation}
\label{sec:res-sem}

Semantic segmentation is a challenge for few-shot learning due to the high intra-class variance of appearance.
For this problem it is crucial to evaluate on not only held-out inputs, but held-out classes, to be certain the few-shot learner has not covertly learned to be an unguided semantic segmentor.
To do so we follow the experimental protocol of \cite{shaban2017one} and score by averaging across four class-wise splits of PASCAL VOC \cite{pascal}, with has 21 classes (including background), and
compare to OSLSM.

Revolver achieves state-of-the-art sparse results that rival the most accurate dense results with just two labeled pixels: see Figure \ref{fig:res-inter-sem}.
OSLSM is incompatible with missing annotations, as it does early fusion by masking, and so is only defined for $\{0,1\}$ annotations.
To evaluate it we map all missing annotations to negative.
Foreground-background is a strong baseline, and we were unable to improve on it with fine-tuning.
The oracle is trained on all classes (nothing is held-out).

For this semantic segmentation problem our method is curiously insensitive to the amount of annotation.
On inspection we determined that the guidance our method extracts has exceptionally low variance for any support in the same class.
This is a pathological solution to optimizing for class generalization from insufficient shot.
Fundamentally, one shot cannot cover the visual variation present in a class.
Consider segmenting a sleek, black dog running given a fluffy, white dog sitting---the guide is forced to be too invariant to color, texture, pose, and more---crippling the few-shot segmentor from improving with shot.

A guided segmentor supervised by instances and trained in the interactive mode behaves differently.
Since the guide must discriminate among instances, it is more sensitive to visual variation.
Learning from \emph{instances} lets our model guide \emph{semantic} tasks from supports that consist of class instances.
Without training on classes and with only one +/- pixel of semantic guidance it is more accurate at $26.4\%$ than OSLSM with the same support, and shows steeper $+6$ improvement at $10$ pixels while the accuracy of our class-supervised segmentor is near constant.
In essence this is an enumerative definition of class as the limit of a collection of instances.

%% file: results_vos.tex
\subsection{Video Object Segmentation}
\label{sec:res-vos}

\begin{figure}
\centering
\begin{subfigure}{0.45\textwidth}
\centering
\includegraphics[width=\textwidth]{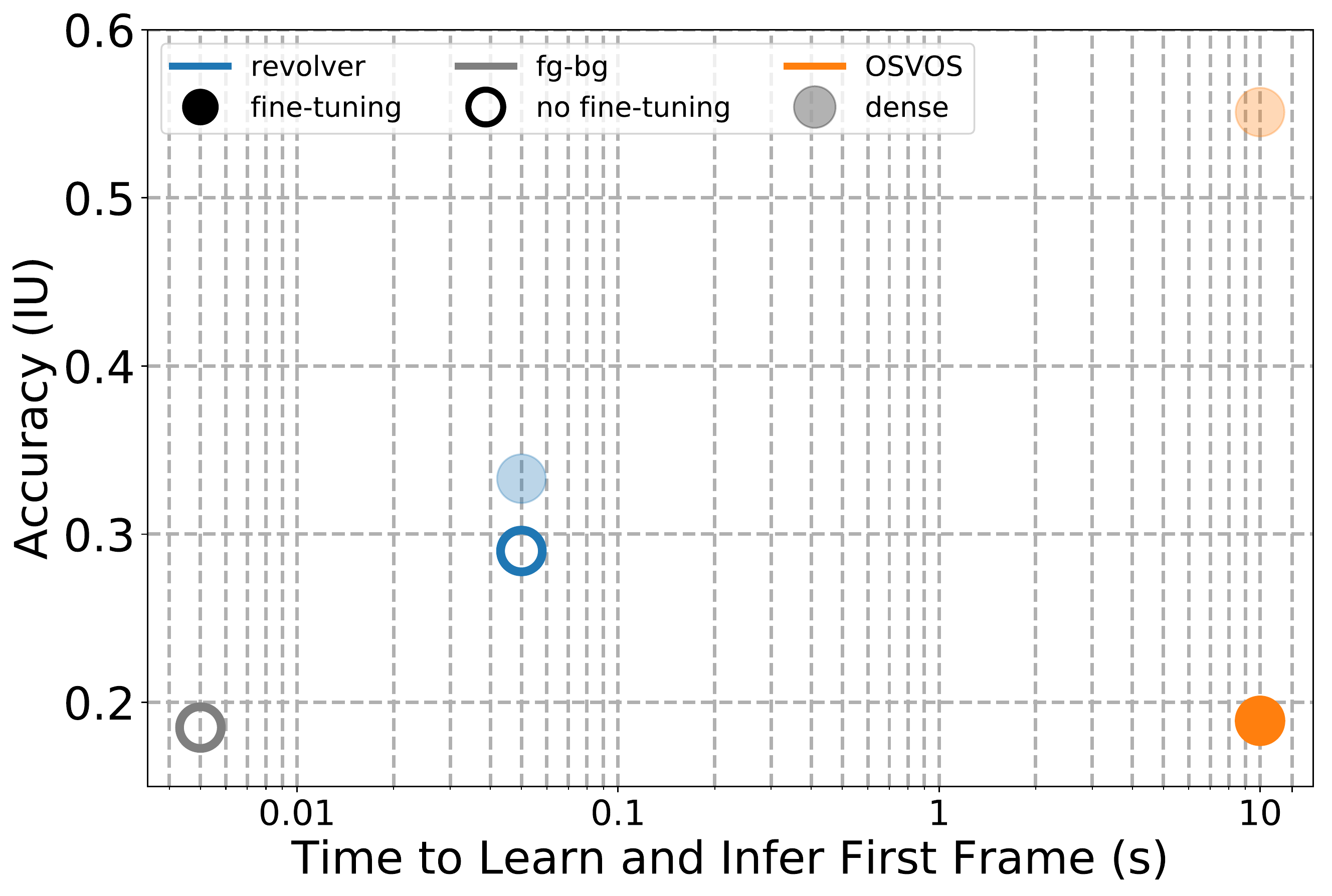}
\end{subfigure}%
\begin{subfigure}{0.45\textwidth}
\centering
\includegraphics[width=\textwidth]{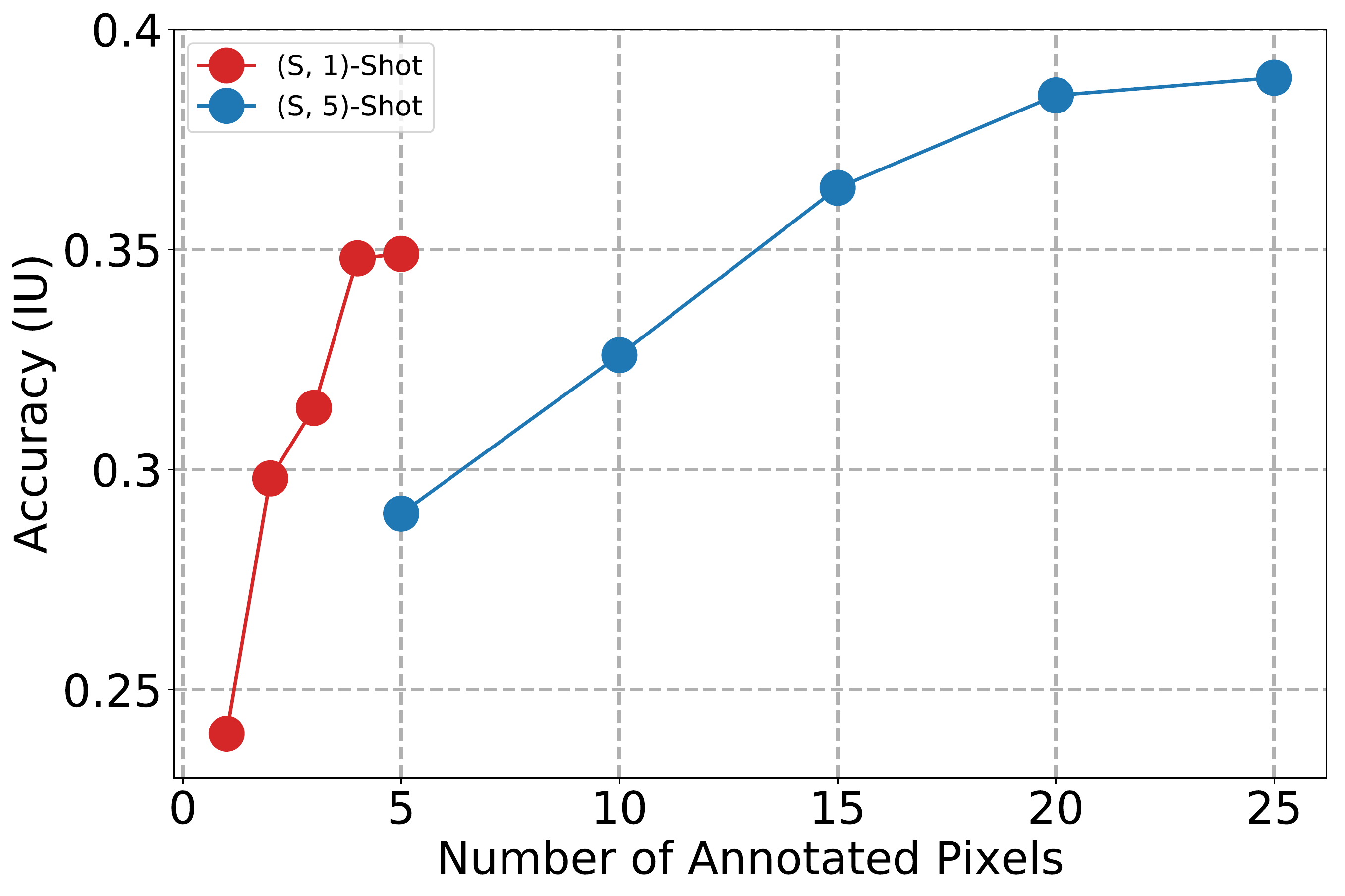}
\end{subfigure}%
\caption{
(left) Accuracy-time evaluation for sparse and dense video object segmentation on DAVIS'17 val.
(right)
Real-time interactive video segmentation on simulated dot interactions.
}
\label{fig:res-vos-iv}
\end{figure}

We evaluate revolver on the DAVIS 2017 benchmark \cite{pont-tuset2017davis} of 2--3 second videos.
For this problem, the object indicated by the fully annotated first frame must be segmented across the video.
We then extend the benchmark to sparse annotations to gauge how methods degrade.
We compare to OSVOS \citep{caelles2016one}, a state-of-the-art method that fine-tunes on the annotated frame and then segments the video frame-by-frame.
Our method is state-of-the-art in accuracy in the sparse regime, and in the dense regime when compared to other approaches of comparable speed.
See Figure \ref{fig:res-vos-iv}.

In the dense regime our method achieves $33.3\%$ accuracy for $80\%$ relative improvement over methods in the same time envelope.
Given (much) more time fine-tuning significantly improves in accuracy, but takes $10+$min/video.
Guidance is $\sim200\times$ faster at 3sec/video.
Our method handles extreme sparsity with little degradation, maintaining $87\%$ of the dense accuracy with only $5$ points for positive and negative.
Fine-tuning struggles to optimize over so few annotations.

%% file: results_intervid.tex
\pagebreak

\subsection{Interactive Video Segmentation}
\label{sec:res-iv}

\begin{wrapfigure}{r}{0.25\textwidth}
\centering
\includegraphics[width=0.25\textwidth]{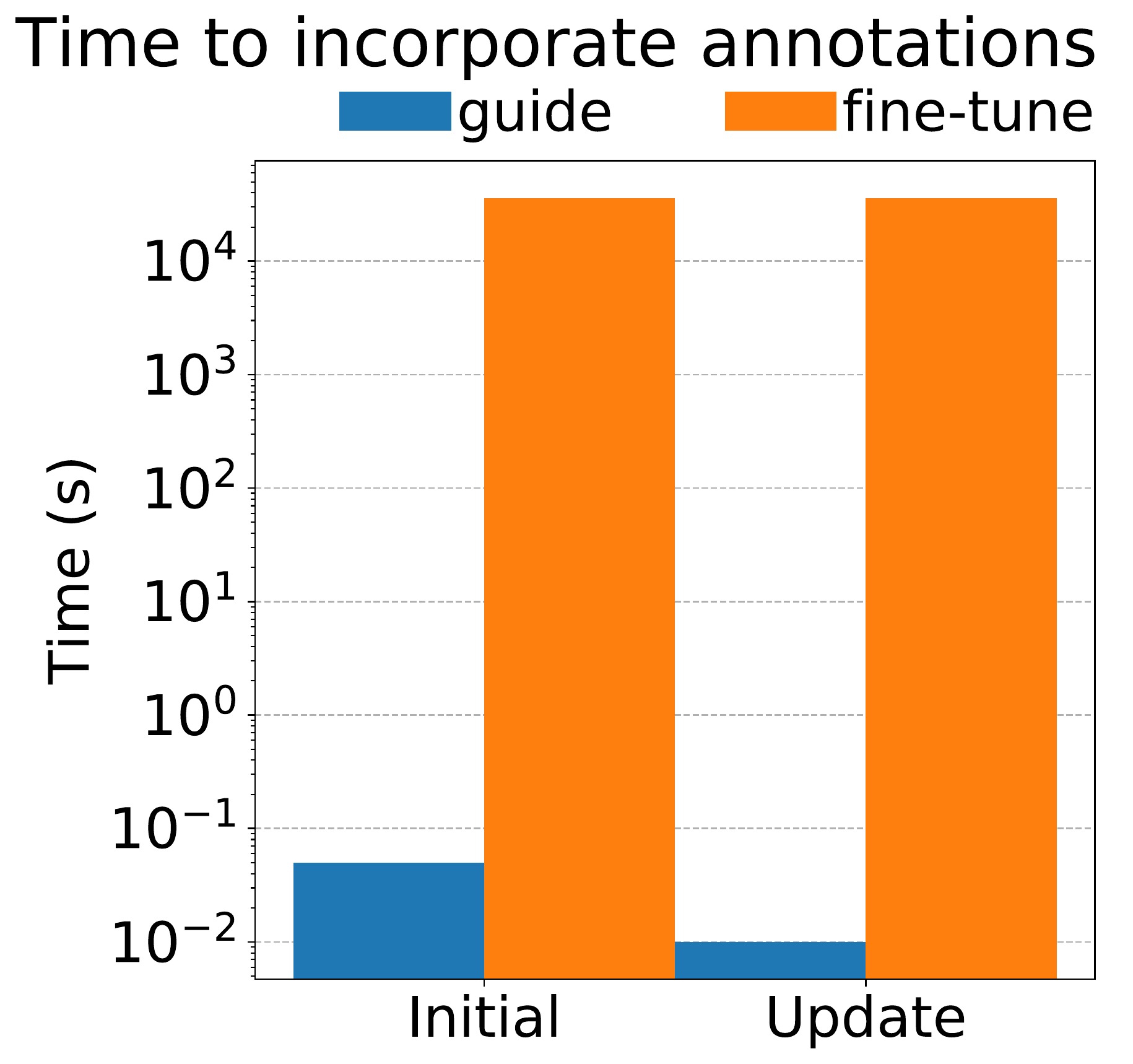}
\end{wrapfigure}

By dividing guidance and inference, revolver can interactively segment video in real time.
As an initial evaluation, we simulate interactions with randomly-sampled dot annotations.
We define a benchmark by fixing the amount of annotation and measuring accuracy as the annotations are given.
The accuracy-annotation tradeoff curve is plotted in Figure \ref{fig:res-vos-iv}.
Our few-shot segmentor improves with both dimensions of shot, whether images ($S$) or pixels ($P$).
Our guided architecture is feedforward and fast, and faster still to update for changes to the annotations (see right).
Existing methods either require annotation on every frame (frame-wise interactive) or are prohibitively slow and cannot handle incremental or sparse annotations (fine-tuning).

%% file: discussion.tex
\section{Discussion}
{Few-shot segmentation unifies annotation-bound segmentation problems. Guided networks reconcile task-driven and interactive inference by extracting guidance, a latent task representation, from any amount of supervision given. With guidance our few-shot segmentor can switch tasks, improve its accuracy near-instantly with more annotations, and segment new inputs without the supervisor.}